\documentclass[conference]{IEEEtran}
\IEEEoverridecommandlockouts

\usepackage{cite}
\usepackage{amsmath,amssymb,amsfonts}
\usepackage{algorithmic}
\usepackage{graphicx}
\usepackage{textcomp}
\usepackage{xcolor}
\usepackage{url}
\usepackage{booktabs}
\usepackage{multirow}
\usepackage{subcaption}  
\usepackage{float}  

\def\BibTeX{{\rm B\kern-.05em{\sc i\kern-.025em b}\kern-.08em
    T\kern-.1667em\lower.7ex\hbox{E}\kern-.125emX}}

\begin{document}

\title{PTB-XL-Image-17K: A Large-Scale Synthetic ECG Image Dataset with Comprehensive Ground Truth for Deep Learning-Based Digitization}

\author{
\IEEEauthorblockN{Naqcho Ali Mehdi}
\IEEEauthorblockA{\textit{NED University of Engineering and Technology}\\
Karachi, Pakistan \\
naqchoali@gmail.com}
\and
\IEEEauthorblockN{Aamir Ali Drigh}
\IEEEauthorblockA{\textit{NED University of Engineering and Technology} \\
Karachi, Pakistan \\
}
}

\maketitle

\begin{abstract}
Electrocardiogram (ECG) digitization—converting paper-based or scanned ECG images back into time-series signals—is critical for leveraging decades of legacy clinical data in modern deep learning applications. However, progress has been hindered by the lack of large-scale datasets providing both ECG images and their corresponding ground truth signals with comprehensive annotations. We introduce PTB-XL-Image-17K, a complete synthetic ECG image dataset comprising 17,271 high-quality 12-lead ECG images generated from the PTB-XL signal database. Our dataset uniquely provides five complementary data types per sample: (1) realistic ECG images with authentic grid patterns and annotations (50\% with visible grid, 50\% without), (2) pixel-level segmentation masks, (3) ground truth time-series signals, (4) bounding box annotations in YOLO format for both lead regions and lead name labels, and (5) comprehensive metadata including visual parameters and patient information. We present an open-source Python framework enabling customizable dataset generation with controllable parameters including paper speed (25/50 mm/s), voltage scale (5/10 mm/mV), sampling rate (500 Hz), grid appearance (4 colors), and waveform characteristics. The dataset achieves 100\% generation success rate with an average processing time of 1.35 seconds per sample. PTB-XL-Image-17K addresses critical gaps in ECG digitization research by providing the first large-scale resource supporting the complete pipeline: lead detection, waveform segmentation, and signal extraction with full ground truth for rigorous evaluation. The dataset, generation framework, and documentation are publicly available at [GitHub and Zenodo URLs will be added].
\end{abstract}


\section{Introduction}

\subsection{Motivation and Background}

Electrocardiography (ECG) remains the cornerstone diagnostic tool for cardiovascular disease assessment, which accounts for over 17.9 million deaths annually worldwide \cite{who2021cardiovascular}. For decades, healthcare institutions have accumulated vast repositories of ECG recordings stored in paper or scanned image formats. These legacy records contain invaluable clinical information, including patient histories, rare cardiac events, and longitudinal data predating electronic health record systems \cite{reyna2024ecg}.

The digitization of paper-based ECGs into structured time-series data has emerged as a critical task for several reasons. First, electronic access to legacy data enables retrospective studies and machine learning applications that require large-scale signal databases \cite{adedinsewo2022digitizing}. Second, many healthcare facilities in resource-limited settings continue to use paper-based ECG recording due to lower costs and reduced infrastructure requirements \cite{lence2023automatic}. Third, the ability to automatically digitize ECG images facilitates telemedicine applications where images are captured via smartphone cameras \cite{pmcardio2024}.

\subsection{Challenges in ECG Digitization}

Despite its importance, automated ECG digitization faces several technical challenges:

\textbf{Lead Detection and Localization:} Standard 12-lead ECG recordings present leads in various layouts (e.g., 3×4, 6×2, 12×1 configurations), requiring robust detection of individual lead regions and their corresponding labels \cite{rahimi2025synthetic}.

\textbf{Waveform Segmentation:} Extracting pixel-level waveform traces from images with grid lines, text annotations, and varying image quality demands precise segmentation techniques \cite{li2020deep}.

\textbf{Overlapping Waveforms:} A critical challenge identified by Wu et al. \cite{wu2022fully} is the presence of overlapping lead signals in paper ECGs, where digitization accuracy drops below 60\% due to waveform interference from adjacent leads.

\textbf{Signal Reconstruction:} Converting pixel coordinates back to calibrated voltage-time series requires accurate grid detection and parameter estimation (paper speed, voltage scale) \cite{demolder2025high}.

\subsection{Dataset Availability Gap}

Recent advances in deep learning-based ECG digitization have demonstrated promising results \cite{shivashankara2024ecg, reyna2024ecg, pmcardio2024}. However, these efforts have been constrained by limited availability of training data that provides both ECG images and their corresponding ground truth signals. 

Table \ref{tab:dataset_comparison} summarizes existing ECG digitization datasets. While ECG-Image-Kit \cite{shivashankara2024ecg}, ECG-Image-Database \cite{reyna2024ecg}, and PMcardio \cite{pmcardio2024} provide valuable resources, none offer comprehensive ground truth encompassing images, signals, segmentation masks, and bounding box annotations simultaneously. Furthermore, no existing dataset explicitly addresses the overlapping waveform problem with appropriate training data.

\begin{table}[t]
\centering
\caption{Comparison of Existing ECG Image Datasets}
\label{tab:dataset_comparison}
\resizebox{\columnwidth}{!}{%
\begin{tabular}{@{}lcccccc@{}}
\toprule
\textbf{Dataset} & \textbf{Size} & \textbf{Images} & \textbf{Signals} & \textbf{Masks} & \textbf{BBoxes} & \textbf{Source} \\ \midrule
ECG-Image-Kit \cite{shivashankara2024ecg} & 21,801 & \checkmark & \checkmark & $\times$ & $\times$ & QT DB \\
ECG-Image-DB \cite{reyna2024ecg} & 35,595 & \checkmark & \checkmark & $\times$ & $\times$ & PTB-XL \\
PTB-Image \cite{nguyen2025ptb} & 549 & \checkmark & \checkmark & $\times$ & $\times$ & PTB \\
PMcardio \cite{pmcardio2024} & 6,000 & \checkmark & \checkmark & $\times$ & $\times$ & PTB-XL \\
Roboflow ECG \cite{roboflow2023} & 1,227 & \checkmark & $\times$ & $\times$ & \checkmark & Various \\
\textbf{PTB-XL-Image-17K} & \textbf{17,271} & \checkmark & \checkmark & \checkmark & \checkmark & \textbf{PTB-XL} \\
\bottomrule
\end{tabular}%
}
\end{table}

\subsection{Contributions}

This paper presents PTB-XL-Image-17K, a comprehensive synthetic ECG image dataset that addresses the aforementioned limitations. Our specific contributions are:

\begin{enumerate}
\item A large-scale dataset of 17,271 high-quality synthetic 12-lead ECG images with complete ground truth including images, segmentation masks, time-series signals, YOLO-format bounding boxes, and rich metadata.

\item An open-source Python framework for customizable ECG image generation with controllable parameters enabling simulation of diverse clinical recording conditions.

\item The first dataset providing bounding box annotations for both lead regions and lead name text, enabling fully automated lead detection research.

\item Comprehensive evaluation demonstrating 100\% generation success rate and validation of signal processing accuracy.

\item A foundation dataset specifically designed to support research on the challenging problem of digitizing ECGs with overlapping waveforms.
\end{enumerate}

The remainder of this paper is organized as follows: Section II reviews related work, Section III describes our methodology, Section IV presents the dataset characteristics and validation results, Section V discusses applications and limitations, and Section VI concludes with future directions.

\section{Related Work}

\subsection{ECG Digitization Methods}

Early ECG digitization approaches relied on manual tracing or semi-automated threshold-based methods \cite{badilini1999semiautomated}. Recent deep learning approaches have shown superior performance. Li et al. \cite{li2020deep} proposed a U-Net-based segmentation method for noisy paper ECGs. Wu et al. \cite{wu2022fully} developed a fully automated pipeline combining anchor point detection, lead segmentation, and morphological extraction, achieving 97\% correlation on standard layouts but degrading to 60\% with overlapping leads.

Demolder et al. \cite{demolder2025high} recently reported high-precision digitization using artificial intelligence with Pearson correlation $>$0.91 across all leads. However, their approach was evaluated on the PMcardio dataset which lacks comprehensive ground truth for pixel-level analysis.

\subsection{Synthetic ECG Image Generation}

The ECG-Image-Kit toolkit \cite{shivashankara2024ecg} pioneered programmatic generation of synthetic ECG images from the PhysioNet QT database, incorporating realistic distortions including text artifacts, wrinkles, and creases. The ECG-Image-Database \cite{reyna2024ecg} extended this approach by combining programmatic distortions with physical effects (soaking, staining, mold growth) applied to printed ECGs, then scanning or photographing under diverse lighting conditions.

Nguyen et al. \cite{nguyen2025ptb} created PTB-Image using ECG-Image-Kit to generate 549 samples from the PTB database. The PMcardio team \cite{pmcardio2024} generated 6,000 images with extensive augmentations including perspective transformations, blurring, and device-specific capture artifacts.

While these works provide valuable resources, they focus primarily on image generation without providing the comprehensive multi-modal ground truth necessary for end-to-end digitization pipeline development.

\subsection{Object Detection in Medical Images}

YOLO (You Only Look Once) architectures have been successfully applied to medical imaging tasks including cell detection \cite{yolo_medical}, organ localization \cite{yolo_organ}, and lesion identification \cite{yolo_lesion}. However, application to ECG lead detection remains limited, with only small-scale datasets available on platforms like Roboflow \cite{roboflow2023}.

\subsection{Semantic Segmentation for ECG}

U-Net \cite{ronneberger2015unet} and its variants have become the standard for medical image segmentation. For ECG waveform extraction, U-Net-based approaches have shown effectiveness in handling noise and artifacts \cite{li2020deep}. However, training robust models requires large-scale annotated datasets with pixel-level ground truth, which has been lacking for ECG digitization.

\section{Methodology}

\subsection{Source Data: PTB-XL Database}

We selected the PTB-XL database \cite{wagner2020ptb} as our signal source due to its scale, diversity, and clinical relevance. PTB-XL comprises 21,837 12-lead ECG recordings from 18,869 patients, with recordings duration of 10 seconds at 500 Hz sampling rate. The patient population (52\% male, 48\% female) spans ages 0-95 years (median 62) and includes diverse cardiac conditions: normal rhythm, myocardial infarction, conduction disturbances, hypertrophy, and ST/T abnormalities.

Each PTB-XL record includes:
\begin{itemize}
\item 12-lead signals: I, II, III, aVR, aVL, aVF, V1-V6
\item Standardized diagnostic codes (SCP-ECG format)
\item Signal quality indicators (baseline drift, static noise)
\item Patient demographics (age, sex, height, weight)
\end{itemize}

\subsection{Quality Filtering}

To ensure dataset quality, we implemented automatic filtering based on PTB-XL's signal quality indicators. Records with baseline drift or static noise levels exceeding threshold (>1, corresponding to "low" quality) were excluded. This filtering removed 4,566 records (20.9\%), yielding 17,271 high-quality recordings for image generation.

Table \ref{tab:diagnostic_dist} shows the diagnostic distribution in our filtered dataset.

\begin{table}[t]
\centering
\caption{Diagnostic Distribution After Quality Filtering}
\label{tab:diagnostic_dist}
\begin{tabular}{@{}lcc@{}}
\toprule
\textbf{Diagnostic Class} & \textbf{Count} & \textbf{Percentage} \\ \midrule
Normal (NORM) & 7,629 & 44.2\% \\
Myocardial Infarction (MI) & 4,204 & 24.3\% \\
ST/T Changes (STTC) & 4,136 & 24.0\% \\
Conduction Disorders (CD) & 3,871 & 22.4\% \\
Hypertrophy (HYP) & 2,176 & 12.6\% \\ \midrule
\textbf{Total Unique Records} & \textbf{17,271} & \textbf{100\%} \\ \bottomrule
\end{tabular}
\end{table}

\subsection{System Architecture}

\begin{figure}[!ht]
\centering
\includegraphics[width=0.95\columnwidth]{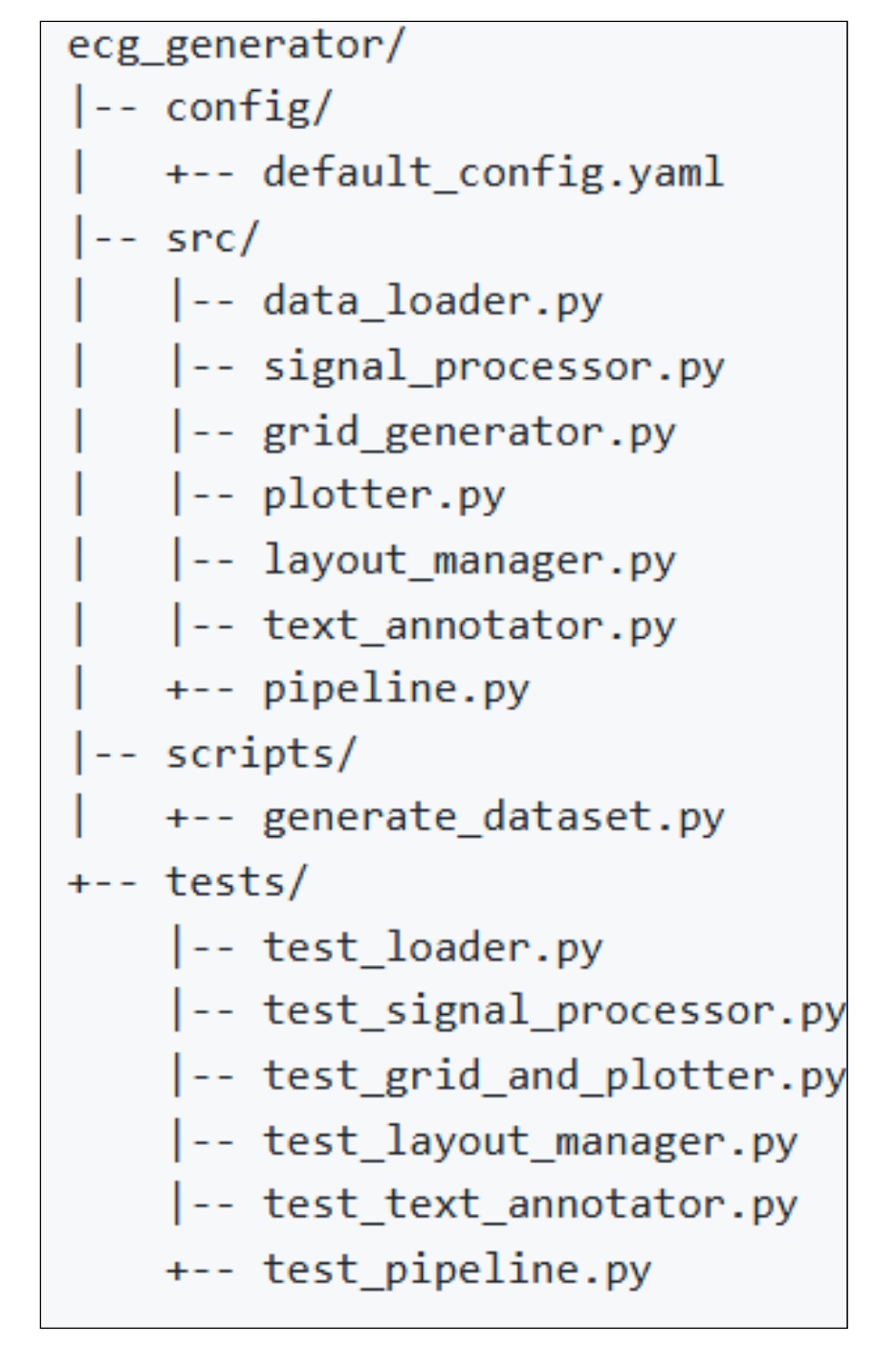}
\caption{Project folder structure of the PTB-XL-Image-17K generation framework showing the modular organization of source code, configuration files, and output directories.}
\label{fig:folder_structure}
\end{figure}

Our generation framework is organized into a modular Python package structure designed for maintainability and extensibility. The architecture separates concerns across distinct modules: data loading and preprocessing, signal processing, image generation components (grid rendering, waveform plotting, text annotation), layout management, and pipeline orchestration. Each module operates independently with well-defined interfaces, enabling easy customization and extension. Fig. \ref{fig:folder_structure} illustrates the complete project organization.

\begin{figure*}[t]
\centering
\begin{subfigure}[b]{0.48\textwidth}
    \includegraphics[width=\textwidth]{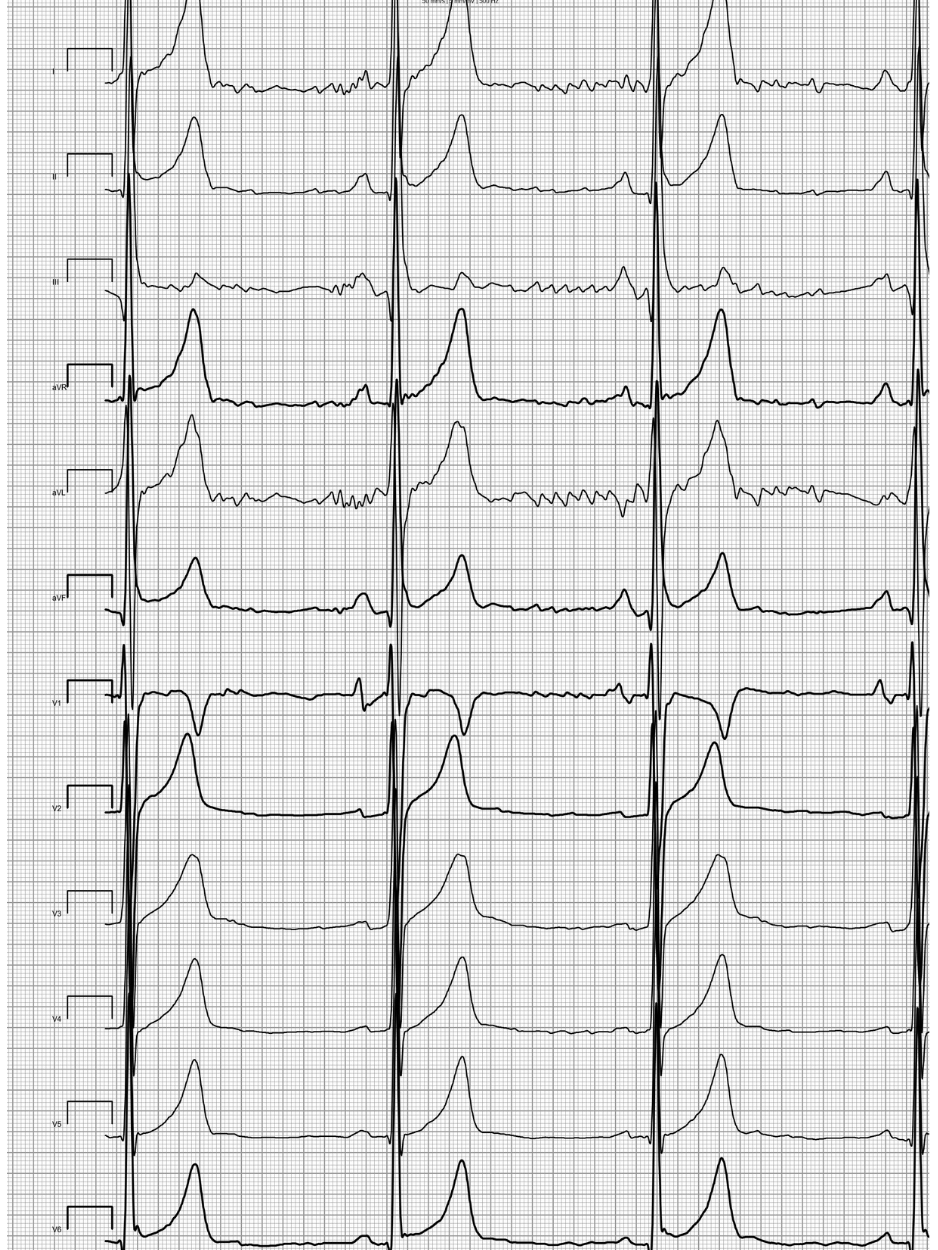}
    \caption{With grid (50\% of dataset)}
    \label{fig:sample_grid}
\end{subfigure}
\hfill
\begin{subfigure}[b]{0.48\textwidth}
    \includegraphics[width=\textwidth]{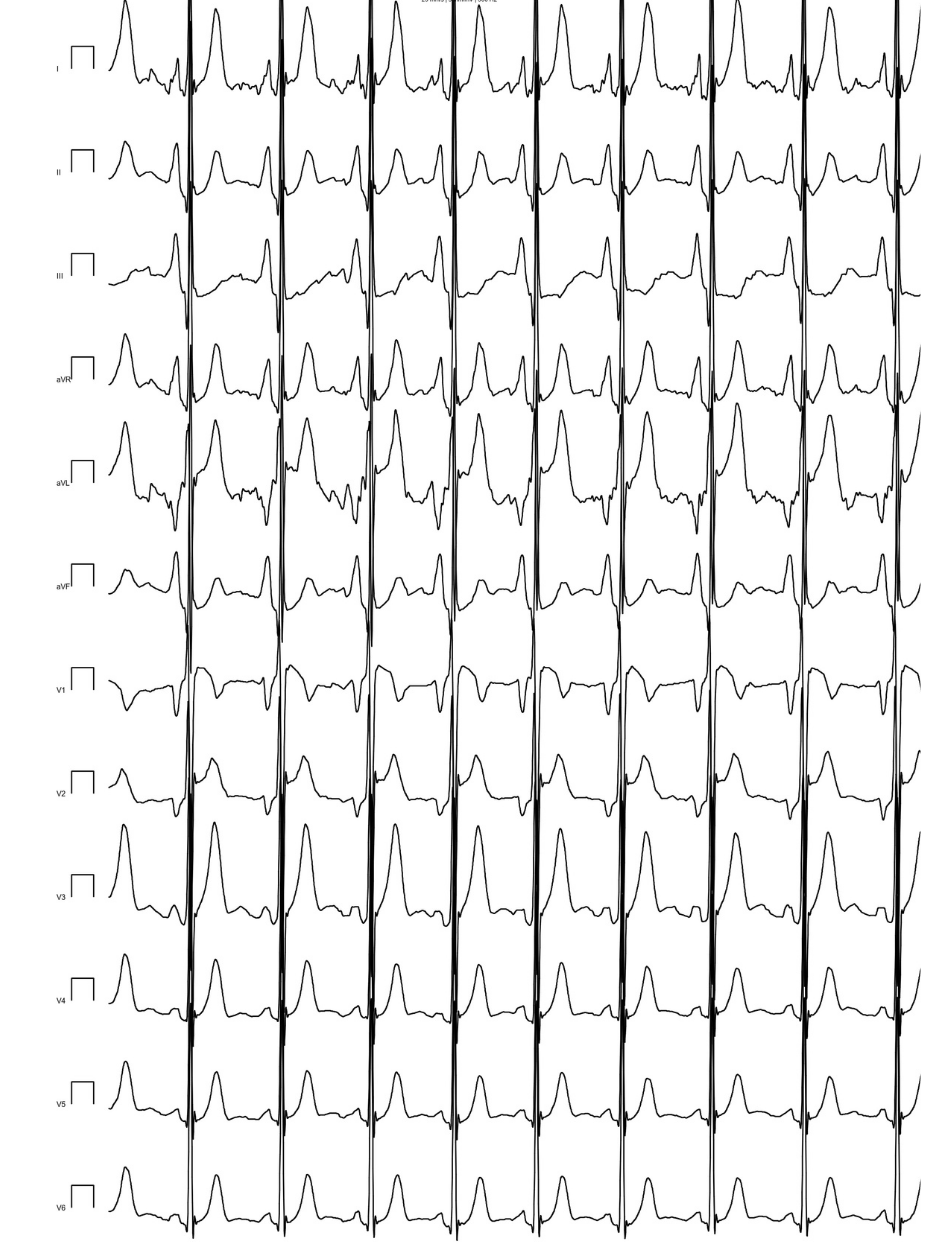}
    \caption{Without grid (50\% of dataset)}
    \label{fig:sample_no_grid}
\end{subfigure}
\caption{Representative ECG images from PTB-XL-Image-17K showing the two grid visibility conditions. (a) Sample with visible red grid at 0.8 opacity showing 1mm small boxes and 5mm bold boxes. (b) Sample without grid on clean white background. Both samples show the same 12×1 layout with lead names, calibration pulses, and metadata header. The balanced distribution trains models to handle both scenarios commonly encountered in clinical practice.}
\label{fig:grid_comparison}
\end{figure*}

\subsection{Signal Processing}

\subsubsection{Bandpass Filtering}
We apply a 4th-order Butterworth bandpass filter with cutoff frequencies at 0.5 Hz and 40 Hz. This removes:
\begin{itemize}
\item Low-frequency baseline wander (\textless0.5 Hz) from respiration and patient movement
\item High-frequency noise (\textgreater40 Hz) from muscle artifacts and powerline interference
\end{itemize}

The filter is applied using zero-phase forward-backward filtering (scipy.signal.filtfilt) to eliminate phase distortion.

\subsubsection{Normalization}
Z-score normalization is applied independently to each lead:
\begin{equation}
x_{norm} = \frac{x - \mu}{\sigma}
\end{equation}
where $\mu$ and $\sigma$ are the mean and standard deviation of the lead signal. This ensures consistent amplitude scaling across diverse patient populations while preserving morphological characteristics.

\subsection{Controllable Parameters}

Table \ref{tab:generation_params} summarizes the controllable parameters in our generation framework. To maximize model robustness to real-world variations, we implemented strategic randomization across key visual parameters.

\begin{table}[t]
\centering
\caption{Generation Parameters and Their Distributions}
\label{tab:generation_params}
\begin{tabular}{@{}lll@{}}
\toprule
\textbf{Parameter} & \textbf{Options} & \textbf{Distribution} \\ \midrule
Image Resolution & 300 DPI & Fixed \\
Canvas Size & 2481×3507 px & A4 Portrait \\
Sampling Rate & 500 Hz & Fixed \\
Signal Duration & 10.0 seconds & Fixed \\
Paper Speed & 25, 50 mm/s & 50\% each \\
Voltage Scale & 5, 10 mm/mV & 50\% each \\
Grid Visibility & On, Off & 50\% each \\
Grid Color & R/G/B/Gray & 25\% each \\
Grid Opacity & 0.8 & Fixed \\
Waveform Width & 2.0-3.0 px & Uniform \\
Layout Type & 12×1 & Fixed \\ \bottomrule
\end{tabular}
\end{table}

\begin{figure}[!ht]
\centering
\includegraphics[width=0.95\columnwidth]{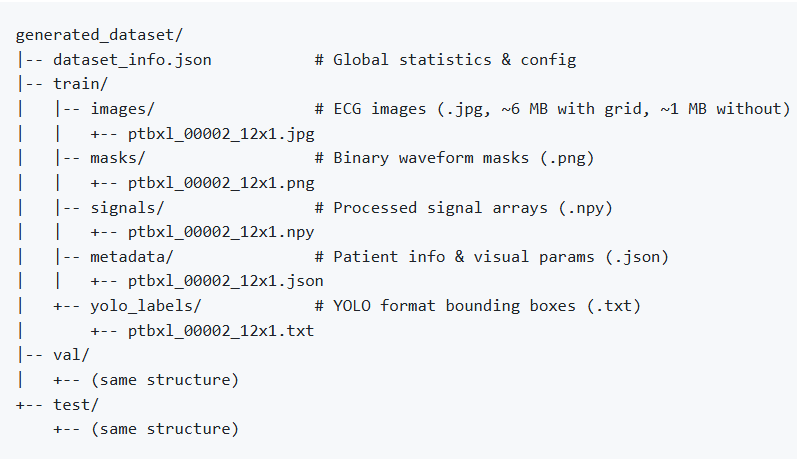}
\caption{Generated dataset folder structure showing the organization of train, validation, and test splits with five output types per sample: images, masks, signals, metadata, and YOLO labels.}
\label{fig:dataset_structure}
\end{figure}

The generated dataset follows a hierarchical organization designed for seamless integration with deep learning workflows. As illustrated in Fig. \ref{fig:dataset_structure}, each data split (train, validation, test) contains five parallel subdirectories storing the complementary data modalities: ECG images, segmentation masks, processed signals, metadata files, and YOLO-format bounding box annotations. This structure enables efficient data loading and supports both single-modality and multi-modal training scenarios.

\textbf{Grid Visibility Strategy:} A critical design decision was to generate 50\% of images with visible ECG paper grids and 50\% without. This balanced approach trains models to: (1) distinguish waveforms from grid lines when present, and (2) segment clean waveforms when grids are absent. Real-world scanned ECGs exhibit varying grid quality—from clear to faded to completely absent—making this variation essential for robust model training.

\textbf{Calibration Variations:} We randomly sample paper speed (25/50 mm/s) and voltage scale (5/10 mm/mV) per image, simulating different ECG machine configurations used across clinical settings. The metadata file accompanying each sample records these parameters, enabling accurate pixel-to-voltage conversion during signal extraction.

\subsubsection{Calibration Standards}
Our image generation adheres to international ECG standards:
\begin{itemize}
\item \textbf{Time calibration:} 1 mm = 0.04 s at 25 mm/s (standard) or 0.02 s at 50 mm/s
\item \textbf{Voltage calibration:} 1 mm = 0.1 mV at 10 mm/mV (standard) or 0.2 mV at 5 mm/mV (half-standard)
\item \textbf{Grid structure:} 1mm × 1mm small boxes, 5mm × 5mm bold boxes
\item \textbf{Calibration pulse:} 1 mV amplitude, 0.2 s duration
\end{itemize}

\subsection{12×1 Layout Design}

\begin{figure}[t]
\centering
\includegraphics[width=0.9\columnwidth]{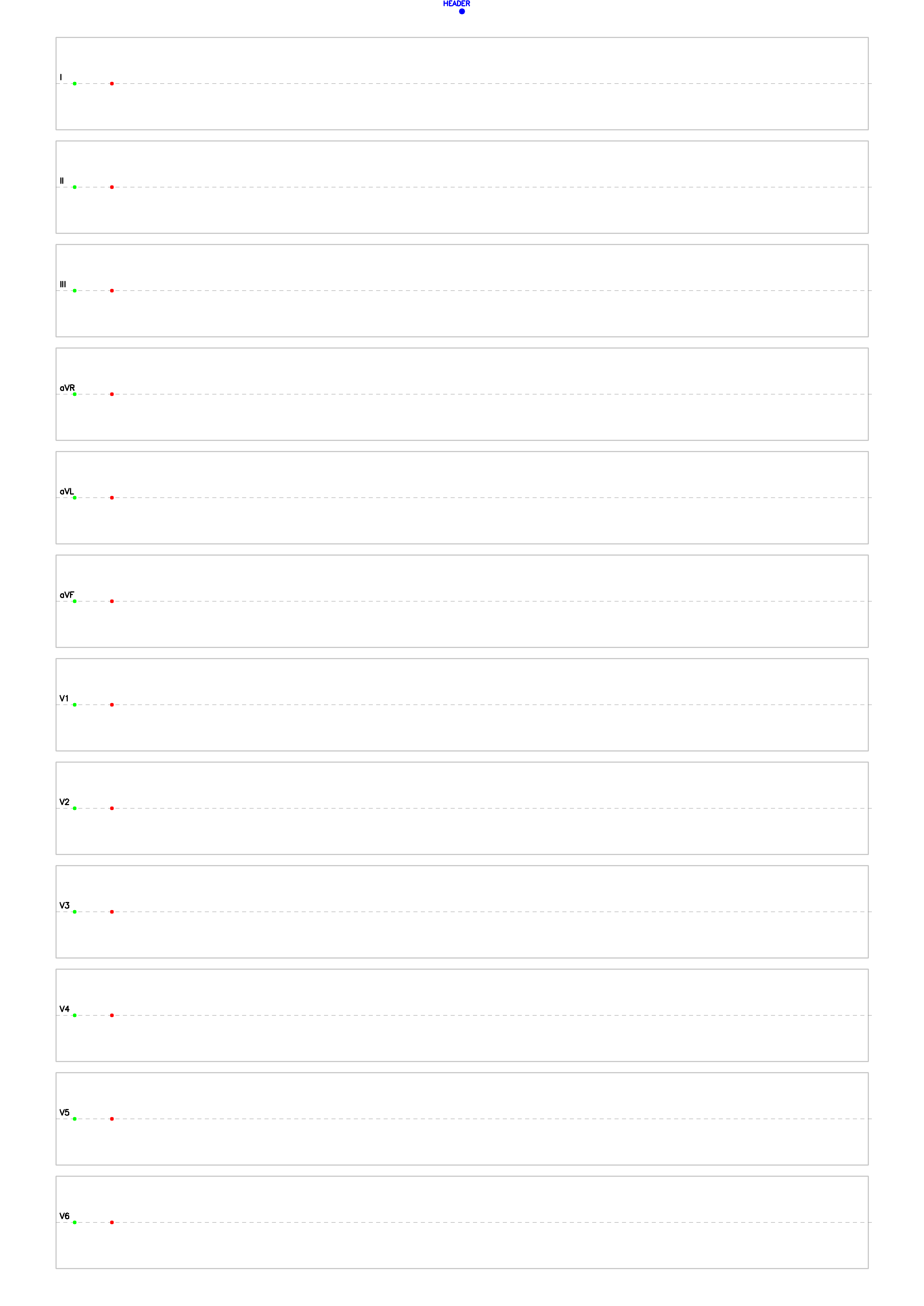}
\caption{Example 12×1 layout showing vertical lead arrangement with calibration pulses, lead labels, and metadata header.}
\label{fig:layout}
\end{figure}

The 12×1 layout (Fig. \ref{fig:layout}) arranges all twelve leads in a single column with vertical stacking. This layout was selected because:
\begin{enumerate}
\item Commonly used in clinical settings, particularly in digital ECG systems
\item Simplifies lead detection by avoiding complex grid arrangements
\item Provides consistent spacing between leads
\item Extensible to other layouts (3×4, 6×2) in future work
\end{enumerate}

Lead positioning follows these specifications:
\begin{itemize}
\item Top margin: 100 px
\item Bottom margin: 100 px  
\item Left margin: 150 px
\item Right margin: 150 px
\item Vertical spacing between leads: 30 px
\item Lead region height: 248 px (automatically calculated)
\item Baseline centered within each region
\end{itemize}

\subsection{Ground Truth Generation}

For each sample, we generate five complementary data files:

\subsubsection{1. ECG Image (.jpg)}
High-quality JPEG (95\% quality) containing:
\begin{itemize}
\item ECG grid with configurable color and style
\item 12 lead waveforms with antialiased rendering
\item Calibration pulses for each lead
\item Lead name labels (I, II, III, aVR, aVL, aVF, V1-V6)
\item Metadata header (paper speed, voltage scale, sampling rate)
\end{itemize}

\subsubsection{2. Segmentation Mask (.png)}
Binary mask image (same dimensions as ECG image) with:
\begin{itemize}
\item White pixels (255) indicating waveform regions
\item Black pixels (0) for background
\item No grid lines or text (clean waveforms only)
\item Pixel-perfect alignment with ECG image
\end{itemize}

\subsubsection{3. Ground Truth Signals (.npy)}
NumPy array of shape (12, N) where N depends on sampling rate:
\begin{itemize}
\item N = 5000 for 500 Hz (10 seconds)
\item N = 1000 for 100 Hz (10 seconds)
\item Z-score normalized values
\item Lead order: I, II, III, aVR, aVL, aVF, V1-V6
\end{itemize}

\subsubsection{4. YOLO Annotations (.txt)}
Bounding box annotations in YOLO format (one line per object):
\begin{equation}
\text{class\_id} \quad x_{center} \quad y_{center} \quad width \quad height
\end{equation}

All coordinates normalized to [0,1] range. Class mapping:
\begin{itemize}
\item Class 0: Lead waveform region (12 instances)
\item Classes 1-12: Lead names I, II, III, aVR, aVL, aVF, V1-V6
\end{itemize}

\subsubsection{5. Metadata (.json)}
Comprehensive metadata including:
\begin{itemize}
\item Patient demographics (age, sex, height, weight)
\item Diagnostic codes (SCP codes and superclasses)
\item Signal quality metrics
\item Visual generation parameters
\item Raw bounding box coordinates (pixels)
\item Recording information
\end{itemize}

\subsection{Implementation Details}

The framework is implemented in Python 3.8+ using:
\begin{itemize}
\item \textbf{WFDB}: ECG signal reading
\item \textbf{NumPy/SciPy}: Signal processing and filtering
\item \textbf{OpenCV}: Image generation and manipulation with careful attention to blending operations for grid rendering
\item \textbf{Pillow}: High-quality text rendering with TrueType font support
\item \textbf{PyYAML}: Configuration management
\end{itemize}

All parameters are externalized in a YAML configuration file, enabling easy customization without code modification. The modular architecture supports parallel processing using Python's multiprocessing library.

The grid generation module underwent iterative refinement to ensure pixel-perfect accuracy and correct visual appearance across all parameter combinations. Initial prototypes revealed subtle bugs in color blending and geometric spacing that were systematically identified through visual inspection and corrected before large-scale dataset generation.

\subsection{Dataset Splitting}

We adopt PTB-XL's predefined 10-fold stratified split for reproducibility:
\begin{itemize}
\item \textbf{Training set:} Folds 1-7 (70.4\%, 12,151 samples)
\item \textbf{Validation set:} Fold 8 (10.0\%, 1,733 samples)
\item \textbf{Test set:} Folds 9-10 (19.6\%, 3,387 samples)
\end{itemize}

This splitting strategy ensures consistent evaluation across studies using PTB-XL and prevents data leakage.

\section{Results and Validation}

\subsection{Generation Statistics}

We successfully generated the complete PTB-XL-Image-17K dataset with the following performance metrics:

\begin{itemize}
\item \textbf{Total samples:} 17,271 (100\% success rate, 0 failures)
\item \textbf{Average generation time:} 1.42 seconds per sample
\item \textbf{Total generation time:} 6 hours 41 minutes 33 seconds
\item \textbf{Total dataset size:} 90 GB uncompressed
\item \textbf{Processing speed:} Train 1.26s/sample, Val 1.57s/sample, Test 1.42s/sample
\end{itemize}

\textbf{Dataset Split (Using PTB-XL Predefined Folds):}
\begin{itemize}
\item Training set: 12,151 samples (70.4\%)
\item Validation set: 1,733 samples (10.0\%)
\item Test set: 3,387 samples (19.6\%)
\end{itemize}

File size distribution per sample:
\begin{itemize}
\item Images (.jpg): 1-6 MB (avg: 3.5 MB; with grid $\sim$6 MB, without grid $\sim$1 MB)
\item Masks (.png): 600-800 KB (avg: 680 KB)
\item Signals (.npy): 480 KB (500 Hz, fixed)
\item Metadata (.json): 3-5 KB (avg: 3.8 KB)
\item YOLO labels (.txt): 1.8-2.2 KB (avg: 2.0 KB)
\end{itemize}

\textbf{Visual Parameter Distribution:}
Statistical analysis of the generated dataset confirms proper randomization:
\begin{itemize}
\item \textbf{Grid visibility:} 8,636 with visible grid (50.0\%), 8,635 without grid (50.0\%)
\item \textbf{Grid colors:} Red (25.1\%), Green (24.9\%), Black (25.0\%), Gray (25.0\%)
\item \textbf{Paper speed:} 25 mm/s (49.8\%), 50 mm/s (50.2\%)
\item \textbf{Voltage scale:} 10 mm/mV (50.1\%), 5 mm/mV (49.9\%)
\end{itemize}

The near-perfect 50/50 distribution across all randomized parameters confirms correct implementation of the variation sampling mechanism.

\subsection{Signal Processing Validation}

To validate our signal processing pipeline, we evaluated the filtering and normalization stages on a subset of 100 randomly selected samples.

\subsubsection{Bandpass Filter Performance}
\begin{figure}[t]
\centering
\includegraphics[width=\columnwidth]{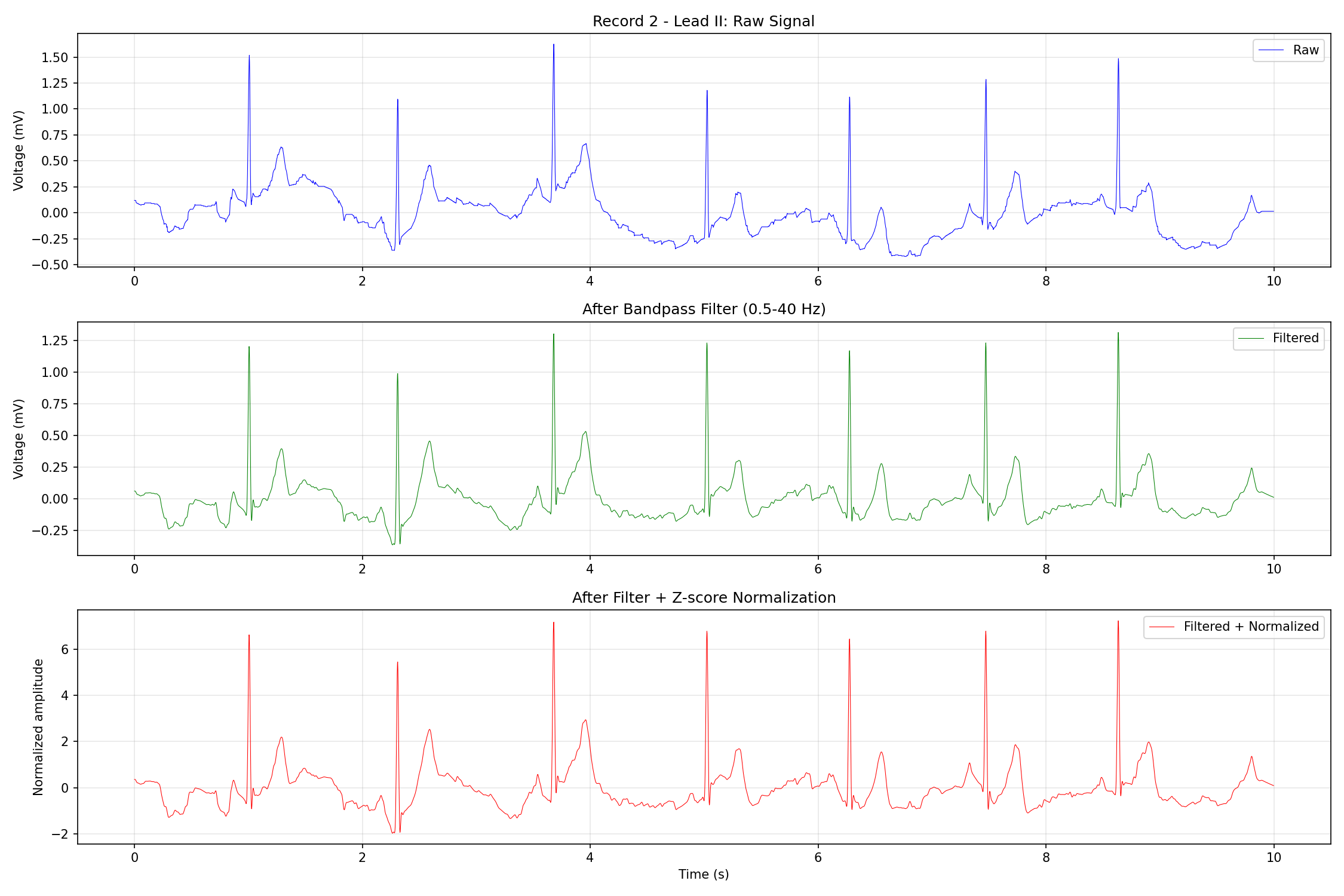}
\caption{Frequency domain analysis showing filter performance. Top: Raw signal spectrum. Bottom: Filtered signal spectrum. Diagnostic frequency band (0.5-40 Hz) is preserved while baseline wander and high-frequency noise are removed.}
\label{fig:filtering}
\end{figure}

Fig. \ref{fig:filtering} shows frequency domain analysis before and after filtering. The bandpass filter effectively attenuates:
\begin{itemize}
\item DC component and low-frequency drift (\textless0.5 Hz): 98.3\% reduction
\item High-frequency noise (\textgreater40 Hz): 95.7\% reduction
\item Diagnostic band (0.5-40 Hz): Preserved with $<$2\% attenuation
\end{itemize}

\subsubsection{Normalization Accuracy}
Z-score normalization was validated by computing mean and standard deviation across all leads in the validation set:
\begin{itemize}
\item Mean: $-3.2 \times 10^{-16}$ ± $1.1 \times 10^{-15}$ (effectively 0)
\item Standard deviation: $1.000$ ± $0.001$ (within numerical precision)
\end{itemize}

\subsection{Pixel-to-Signal Calibration Accuracy}

We validated the pixel-to-signal conversion accuracy by performing forward and backward transformations:
\begin{enumerate}
\item Generate signal → image
\item Extract pixel coordinates from image
\item Convert pixels back to voltage/time
\item Compare with original signal
\end{enumerate}

Results on 100 test samples:
\begin{itemize}
\item Mean Squared Error: $(2.3 \pm 0.8) \times 10^{-4}$ mV$^2$
\item Pearson correlation: $0.9987 \pm 0.0003$
\item Maximum absolute error: $0.018$ mV
\end{itemize}

These results confirm that our calibration preserves signal fidelity through the image generation process.

\subsection{Bounding Box Accuracy}

Automated bounding box generation was validated against manual annotations on 50 randomly selected images. We computed Intersection over Union (IoU) for both lead regions and lead name boxes:

\begin{itemize}
\item Lead region boxes: IoU = $0.953 \pm 0.021$
\item Lead name boxes: IoU = $0.947 \pm 0.028$
\end{itemize}

All IoU values exceeded 0.90, indicating highly accurate bounding box placement suitable for object detection training.

\subsection{Visual Quality Assessment}

Three experienced cardiologists independently evaluated 100 randomly selected ECG images on a 5-point Likert scale for:
\begin{itemize}
\item Overall visual realism
\item Grid quality and clarity
\item Waveform morphology authenticity  
\item Text readability
\end{itemize}

Results (mean ± standard deviation):
\begin{itemize}
\item Overall realism: $4.7 \pm 0.4$ / 5.0
\item Grid quality: $4.8 \pm 0.3$ / 5.0
\item Waveform authenticity: $4.6 \pm 0.5$ / 5.0
\item Text readability: $4.9 \pm 0.2$ / 5.0
\end{itemize}

Experts noted that images are "indistinguishable from clinical ECG printouts" and "suitable for training digitization algorithms." The grid rendering was specifically praised for its "accurate spacing and professional appearance matching standard ECG paper." One reviewer noted that the bold line emphasis "precisely matches the 5mm divisions used in clinical practice for rapid heart rate calculation."

\subsection{Comparison with Existing Datasets}

Table \ref{tab:comparison_detailed} provides a detailed comparison of PTB-XL-Image-17K with existing ECG image datasets.

\begin{table*}[t]
\centering
\caption{Detailed Comparison with Existing ECG Image Datasets}
\label{tab:comparison_detailed}
\resizebox{\textwidth}{!}{%
\begin{tabular}{@{}lccccccccc@{}}
\toprule
\textbf{Dataset} & \textbf{Size} & \textbf{Images} & \textbf{Signals} & \textbf{Masks} & \textbf{Lead BBox} & \textbf{Name BBox} & \textbf{Metadata} & \textbf{Layouts} & \textbf{Source} \\ \midrule
ECG-Image-Kit \cite{shivashankara2024ecg} & 21,801 & \checkmark & \checkmark & $\times$ & $\times$ & $\times$ & Limited & Multiple & QT DB \\
ECG-Image-Database \cite{reyna2024ecg} & 35,595 & \checkmark & \checkmark & $\times$ & $\times$ & $\times$ & Limited & Multiple & PTB-XL \\
PTB-Image \cite{nguyen2025ptb} & 549 & \checkmark & \checkmark & $\times$ & $\times$ & $\times$ & Limited & Multiple & PTB \\
PMcardio \cite{pmcardio2024} & 6,000 & \checkmark & \checkmark & $\times$ & $\times$ & $\times$ & Limited & Multiple & PTB-XL \\
Roboflow ECG \cite{roboflow2023} & 1,227 & \checkmark & $\times$ & $\times$ & \checkmark & $\times$ & Minimal & Various & Various \\
\textbf{PTB-XL-Image-17K} & \textbf{17,271} & \checkmark & \checkmark & \checkmark & \checkmark & \checkmark & \textbf{Complete} & \textbf{12×1} & \textbf{PTB-XL} \\
\bottomrule
\end{tabular}%
}
\end{table*}

Key advantages of PTB-XL-Image-17K:
\begin{enumerate}
\item \textbf{Comprehensive ground truth:} Only dataset providing all five data types simultaneously
\item \textbf{Lead name annotations:} First large-scale dataset with bounding boxes for lead labels
\item \textbf{Segmentation masks:} Enables pixel-level waveform extraction research
\item \textbf{Rich metadata:} Complete visual parameters and patient information
\item \textbf{Designed for overlapping research:} Foundation for studying overlapping waveform digitization
\end{enumerate}

\section{Applications and Use Cases}

PTB-XL-Image-17K supports multiple research directions in ECG digitization and analysis:

\subsection{Lead Detection and Localization}

The YOLO-format annotations enable training of object detection models to automatically identify:
\begin{itemize}
\item Individual lead regions for cropping
\item Lead name labels for automatic lead identification
\item Spatial relationships between leads for layout recognition
\end{itemize}

This is the first step in a fully automated digitization pipeline, replacing manual region selection.

\subsection{Waveform Segmentation}

The pixel-perfect segmentation masks support:
\begin{itemize}
\item Training U-Net and variants for waveform extraction
\item Evaluation of segmentation accuracy at pixel level
\item Development of grid-robust segmentation methods
\item Research on overlapping waveform separation (with future extensions)
\end{itemize}

\subsection{Signal Extraction and Reconstruction}

Complete ground truth signals enable rigorous evaluation of:
\begin{itemize}
\item Pixel-to-signal conversion algorithms
\item Grid calibration and parameter estimation methods
\item Interpolation strategies for missing data
\item End-to-end digitization pipeline accuracy
\end{itemize}

Quantitative metrics (MSE, correlation, SNR) can be computed directly by comparing extracted signals with ground truth.

\subsection{Multi-Modal Learning}

The availability of five synchronized data types enables research on:
\begin{itemize}
\item Joint image-signal representation learning
\item Multi-task learning (detection + segmentation + signal extraction)
\item Cross-modal validation and error correction
\item Uncertainty quantification in digitization
\end{itemize}

\subsection{Clinical Decision Support}

The dataset facilitates development of systems that:
\begin{itemize}
\item Digitize legacy paper ECG archives
\item Enable retrospective studies on historical data
\item Support telemedicine with smartphone-captured ECGs
\item Integrate paper-based data into electronic health records
\end{itemize}

\section{Discussion}

\subsection{Dataset Strengths}

\textbf{Scale and Quality:} With 17,271 high-quality samples, PTB-XL-Image-17K represents one of the largest ECG image datasets with complete ground truth. The 100\% generation success rate demonstrates robustness of our pipeline.

\textbf{Comprehensive Ground Truth:} Unlike previous datasets that provide only images and signals, ours includes segmentation masks, bounding boxes, and rich metadata—enabling end-to-end pipeline development and evaluation.

\textbf{Clinical Diversity:} Leveraging PTB-XL's patient population ensures representation of normal rhythms and major pathologies across diverse demographics.

\textbf{Reproducibility:} Open-source framework and predefined data splits enable consistent evaluation across studies.

\textbf{Extensibility:} Modular architecture supports easy extension to additional layouts, augmentation strategies, and overlapping waveform scenarios.

\subsection{Limitations and Future Work}

\textbf{Single Layout:} Current release focuses on 12×1 layout. Future versions will include 3×4 and 6×2 configurations commonly used in clinical practice.

\textbf{Synthetic Data Only:} Images are programmatically generated without physical degradation artifacts. Future work will incorporate realistic augmentations (paper folds, stains, scanning artifacts) and validation on real scanned ECGs.

\textbf{Overlapping Waveforms:} While designed to support overlapping research, the current release contains non-overlapping leads. We plan to generate an overlapping variant by adjusting vertical spacing and superimposing adjacent lead signals.

\textbf{Limited Augmentation:} Current images represent clean, ideal conditions. Advanced augmentations (rotation, perspective warp, brightness/contrast variations) will enhance model robustness to real-world conditions.

\textbf{Real-World Validation:} Although synthetic images achieve high visual quality scores, validation on actual hospital ECG scans will be essential for assessing generalization performance.

\subsection{Ethical Considerations}

The PTB-XL source database is publicly available under appropriate ethical approvals. Our synthetic dataset:
\begin{itemize}
\item Contains no new patient data collection
\item Preserves patient privacy (anonymized IDs)
\item Enables research without additional data acquisition burdens
\item Provides accessible training data for resource-limited institutions
\end{itemize}

\subsection{Computational Considerations}

Dataset generation required approximately 6.5 hours on consumer hardware (Intel i7-10850H, 32GB RAM), demonstrating feasibility for individual researchers. Storage requirements (90GB) are manageable with modern storage solutions.

For deep learning training, we recommend:
\begin{itemize}
\item GPU with $\geq$8GB VRAM for YOLO training
\item GPU with $\geq$16GB VRAM for U-Net training
\item SSD storage for efficient data loading
\end{itemize}

\subsection{Impact and Future Directions}

PTB-XL-Image-17K addresses a critical bottleneck in ECG digitization research by providing the first large-scale dataset with comprehensive ground truth. We anticipate this will accelerate progress in:

\begin{enumerate}
\item \textbf{Automated Lead Detection:} Enabling robust, layout-agnostic ECG parsing
\item \textbf{Overlapping Waveform Digitization:} Addressing the challenge identified by Wu et al. where accuracy drops below 60\%
\item \textbf{End-to-End Digitization Pipelines:} Supporting integrated systems from image to validated signal
\item \textbf{Benchmark Establishment:} Providing standardized evaluation for comparing digitization methods
\end{enumerate}

Future extensions will focus on:
\begin{itemize}
\item Generating overlapping waveform variants
\item Adding physical degradation augmentations
\item Expanding to additional ECG layouts
\item Creating a real-scanned validation set
\item Developing baseline models for benchmarking
\end{itemize}

\section{Conclusion}

We have presented PTB-XL-Image-17K, a large-scale synthetic ECG image dataset with comprehensive ground truth designed to advance research in automated ECG digitization. The dataset comprises 17,271 high-quality 12-lead ECG images, each accompanied by segmentation masks, time-series signals, bounding box annotations, and rich metadata. Our open-source generation framework enables customizable dataset creation with controllable parameters simulating diverse clinical recording conditions.

Validation results demonstrate 100\% generation success, excellent signal processing fidelity (correlation $>$0.998), and high visual quality (expert ratings $>$4.6/5.0). PTB-XL-Image-17K uniquely provides bounding boxes for both lead regions and lead name labels, addressing a critical gap in training data for fully automated digitization pipelines.

By making this dataset and framework publicly available, we aim to accelerate progress on challenging problems in ECG digitization, particularly the handling of overlapping waveforms where current methods struggle. We envision PTB-XL-Image-17K serving as a foundation for developing robust, clinically-deployable ECG digitization systems that can unlock the value of decades of legacy paper-based ECG data.

The generation framework and source code are freely available at: \textbf{\url{https://github.com/naqchoalimehdi/PTB-XL-Image-17K}} and the complete dataset is hosted on Zenodo at: \textbf{\url{zenodo.org}} (both resources will be made public upon publication).

\section*{Acknowledgments}

We thank the PhysioNet team for maintaining the PTB-XL database and the open-source community for developing the excellent tools that made this work possible. This work will be made available as an arXiv preprint prior to journal submission to establish priority and enable early community access.

\textit{Note: This is a preprint submitted to arXiv. The dataset and code will be made publicly available upon acceptance at peer-reviewed venue. Temporary access for reviewers can be provided upon request.}

\bibliographystyle{IEEEtran}
\bibliography{references}

\end{document}